\documentclass[
twocolumn,
]{ceurart}

\usepackage{listings}

\usepackage[american]{babel}

\usepackage{mathtools} 
\usepackage{booktabs} 
\usepackage{tikz} 
\usepackage{amsmath}
\usepackage{amssymb}
\usepackage{tikz}
\usepackage{rotating}
\usepackage{graphicx}
\usetikzlibrary{bayesnet}
\usepackage{cleveref}
\usepackage{bm}
\usepackage{wrapfig}
\DeclareMathOperator*{\argmax}{argmax} 
\usepackage{standalone}
\usepackage{natbib}
\usepackage{xcolor}
\usepackage{algorithm}
\usepackage{comment}
\usepackage{wrapfig}
\usepackage{float} 

\usepackage{algpseudocode}

\begin{document}

\copyrightyear{2022}
\copyrightclause{Copyright for this paper by its authors.
  Use permitted under Creative Commons License Attribution 4.0
  International (CC BY 4.0).}

\conference{BMAW 2022: 16th Bayesian Modelling Applications Workshop}

\title{
Kernel Learning for Explainable Climate Science}


\author[1]{Vidhi Lalchand}
\address[1]{Department of Physics, University of Cambridge, Cambridge, UK}

\author[2,3]{Kenza Tazi}[%
email= kt484@cam.ac.uk]
\address[2]{Department of Engineering, University of Cambridge, Cambridge, UK}
\cormark[1]

\author[2]{Talay Cheema}
\author[2]{Richard Turner}

\author[3]{Scott Hosking}
\address[3]{British Antarctic Survey, Cambridge, UK}

\cortext[1]{Corresponding author.}

\begin{abstract}
The Upper Indus Basin, Himalayas provides water for 270 million people and countless ecosystems. However, precipitation, a key component to hydrological modelling, is poorly understood in this area. A key challenge surrounding this uncertainty comes from the complex spatial-temporal distribution of precipitation across the basin. In this work we propose Gaussian processes with structured non-stationary kernels to model precipitation patterns in the UIB. 
Previous attempts to quantify or model precipitation in the Hindu Kush Karakoram Himalayan region have often been qualitative or include crude assumptions and simplifications which cannot be resolved at lower resolutions. This body of research also provides little to no error propagation. We account for the spatial variation in precipitation with a non-stationary Gibbs kernel parameterised with an input dependent lengthscale. This allows the posterior function samples to adapt to the varying precipitation patterns inherent in the distinct underlying topography of the Indus region. The input dependent lengthscale is governed by a latent Gaussian process with a stationary squared-exponential kernel to allow the function level hyperparameters to vary smoothly. In ablation experiments we motivate each component of the proposed kernel by demonstrating its ability to model the spatial covariance, temporal structure and joint spatio-temporal reconstruction. We benchmark our model with a stationary Gaussian process and a Deep Gaussian processes.

\end{abstract}


\maketitle

\section{Motivation}

The Indus River is one of the longest rivers in Asia, sustaining the livelihoods of over 268 million people \citep{wester2019hindu}. The river and its tributaries provide fresh water for drinking, domestic usage, industrial processes, and agriculture through the world’s largest contiguous irrigation system \citep{basharat2019water}. The river also delivers most of Pakistan’s electricity through hydropower plants \citep{nie2021glacial} and supports countless ecosystems and biodiversity hotspots \citep{xu2019sustaining}.

Of all the rivers originating in the Himalayas, the Indus depends most strongly on water from snow and glacier found in the Upper Indus Basin (UIB) \citep{lutz2014consistent}. Over 60\% of the Indus’ annual flow is attributed to the springtime melt of the snowpack and glaciers \citep{immerzeel2010climate}. As climate change progresses, these drivers are expected to be replaced by precipitation. This change will lead to on average less but more extreme variations in river flow and in turn more floods, landslides, and droughts \citep{huss2017toward}.
 
With no robust adaption measures, financial losses and profound socio-economic consequences including food and water scarcity, mass migration and violent conflict are projected \citep{wester2019hindu, huss2017toward}. The scale of these ramifications is still unknown with the largest source of uncertainty attributed to precipitation. \citep{li2016does, wulf2016differentiating, remesan2015effect, meng2014statistical, andermann2011evaluation}.

\begin{figure*}[t]
    \includegraphics[scale=0.70]{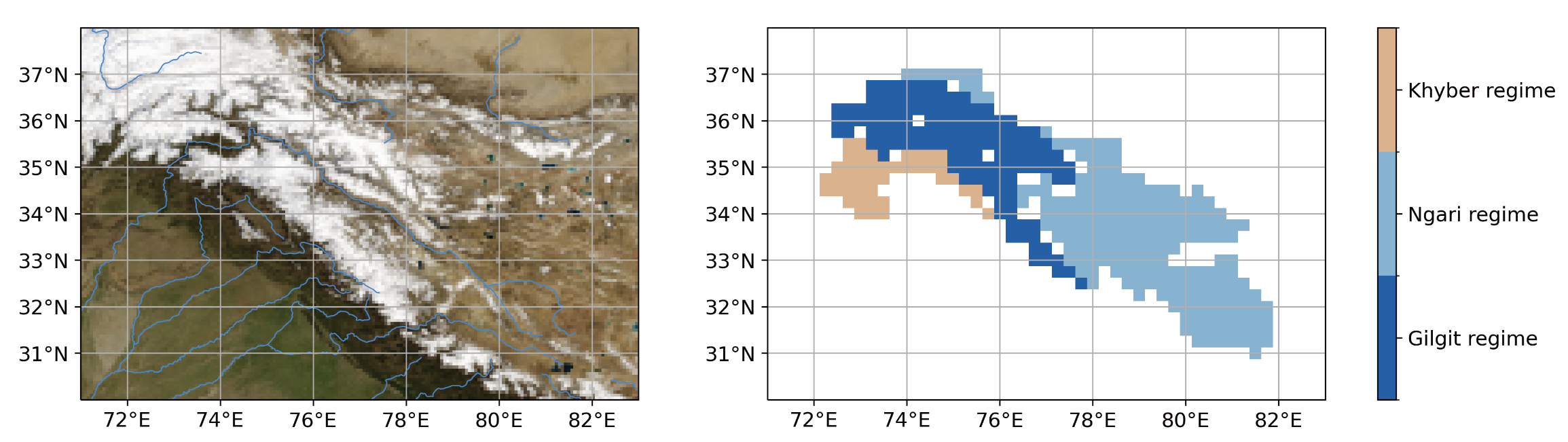}
    \caption{Left: Shaded-relief image of the topography of the Upper Indus Basin with natural water bodies overlaid. Right: Three precipitation regimes in the UIB identified through kmeans clustering. Regime names were chosen by the authors.}
    \label{fig:map_and_regimes}
\end{figure*}

Previous research modelling and predicting precipitation in the Hindu Kush Karakoram Himalayan region are often qualitative or include assumptions and simplifications which cannot be resolved at lower
resolutions \citep{dahri2016appraisal}. For example, the regional climates models from Coordinated Regional Climate Downscaling Experiment for South Asia regularly overestimates historical precipitation in the Himalayas by over 100\% for both winter and summer \citep{sanjay2017downscaled}. These models are also computationally expensive, lack error propagation, and generate large model-dependent uncertainty.

Gaussian processes (GPs) offer a versatile and interpretable way of studying and predicting precipitation in this area. Crucially, they offer two unique advantages which preclude several other modelling tools. First, GP estimates provide principled uncertainty quantification by design. One can derive concrete prediction intervals through the posterior predictive distribution; predictions accompanied by robust uncertainty estimates can be critical for downstream decision making. Secondly, GPs are a flexible prior of functions and allow one to encode specific inductive biases through kernel construction. For instance, one can encode properties like periodicity, smoothness or spatial heterogeneity through careful specification of the covariance kernel. 

This work concerns interpretable kernel constructions for precipitation modelling in the UIB. We account for non-stationarity and spatial heterogeneity through a spatio-temporal kernel. In the next section, we give a brief overview of the GP framework along with non-stationary kernels. \Cref{cs} presents experimental results from the case-study where we leverage spatio-temporal and non-stationary kernels.

\section{Background}
\label{back}

GPs are a powerful probabilistic and non-parametric tool for modelling functions. They are fully specified by a mean and covariance function where the latter controls the inductive bias and support of functions under the prior. The choice of the covariance function (alternatively, kernel or kernel function) and in turn selecting the hyperparameters of the covariance function is jointly referred to as the \textit{model selection} problem \citep{rasmussen2005gaussian} in GPs. A large cross-section of Gaussian process literature uses universal kernels like the squared exponential (SE) kernel along with automatic relevance determination (ARD) in high-dimensions. The SE-ARD kernel is a translation-invariant stationary kernel which gives infinitely smooth and differentiable samples in function space. The ARD framework operates by pruning away extraneous dimensions through contracting their inverse-lengthscales. The SE-ARD framework is the most commonly reported baseline for Gaussian process regression tasks. The SE-ARD kernel is given by, 
\begin{equation}
    k(\bm{x}, \bm{x}^{\prime})_{\text{SE-ARD}} = 
    \sigma^{2}_{f}\exp \Bigg\{-\dfrac{1}{2}\sum_{d=1}^{D}\dfrac{(x_{d} - x^{\prime}_{d})^{2}}{\ell_{d}^{2}} \Bigg\} 
\end{equation}

where $\bm{x}, \bm{x}^{\prime} \in X$ are high-dimensional inputs $X \equiv \{\bm{x}_{i}\}_{i=1}^{N}$, each $\bm{x}_{i} \in \mathbb{R}^{D}$ and $\{\ell_{d}\}_{d=1}^{D}$ denotes a scalar lengthscale per dimension and $\sigma^{2}_{f}$ is usually a scalar amplitude.

 Given observations $(X, \bm{y}) = \lbrace{ \bm{x}_{i}, y_{i} \rbrace}_{i=1}^{N}$  where $y_{i}$ are noisy realizations of some latent function values $f$ corrupted with Gaussian noise, $y_{i} = f(\bm{x}_{i}) + \epsilon_{i}$, $\epsilon_{i} \in \mathcal{N}(0, \sigma_{n}^{2})$, let $k_{\theta}(\bm{x}_{i}, \bm{x}_{j})$ denote a positive definite covariance function parameterised with hyperparameters $\theta$. The generative model governing the data is given by, %
 \begin{align}
     f &\sim \mathcal{GP}(0, k_{\theta}(.,.)) \nonumber \\
     p(\bm{y}|f) &= p(y_{i} | f_{i}) = \prod_{i=1}^{N}\mathcal{N}(f_{i}, \sigma^{2}_{n})
 \end{align}%
 
 A Gaussian noise setting which we assume through out yields a closed-form marginal likelihood $p(\bm{y}|\theta)$. In the standard set-up the negative log marginal likelihood serves as the loss function against which $\theta$ is optimised, we call this procedure ML-II. The SE-ARD kernel with ML-II inference serves as the baseline model to benchmark performance for the variants proposed in this work. Learning occurs through adaptation of the hyperparameters ($\theta$) of the covariance function. In the case of the SE-ARD kernel described above, $\theta = \{\sigma^{2}_{f}, \{\ell_{d}\}_{d=1}^{D}, \sigma^{2}_{n}\}$.  
 \begin{figure*}[t]
    \centering
    \includegraphics[scale=0.65]{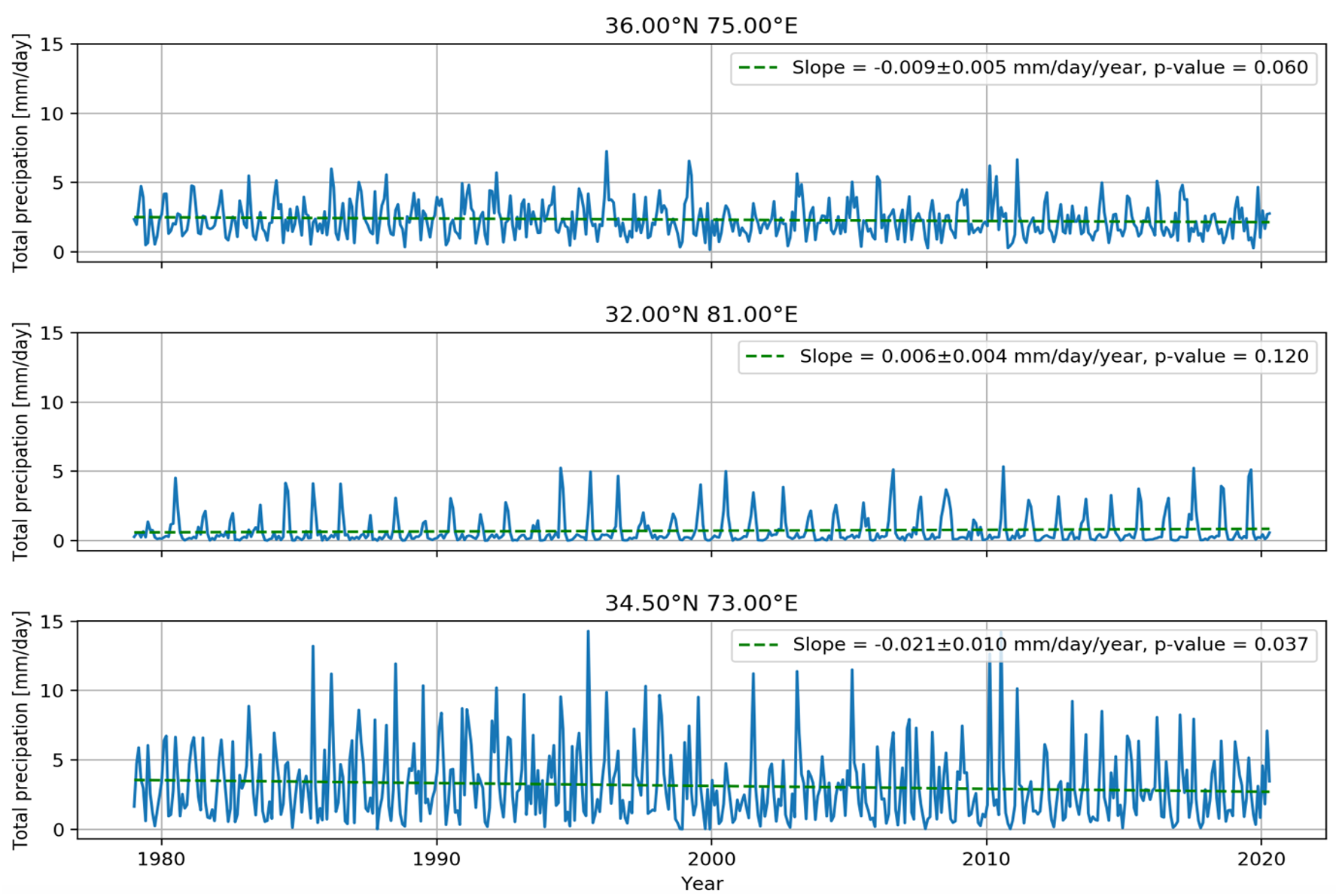}
    \caption{Timeseries drawn from each precipitation regime shown in Figure \ref{fig:map_and_regimes} (top: Gilgit, middle: Ngari, bottom: Khyber). Note that there are no obvious trends in average or extreme precipitation during this period.}
    \label{fig:typical}
\end{figure*}
\begin{align}
p(\bm{y}|\theta) &= \int p(\bm{y}|f)p(\bm{f}|\theta)df \\
&= \int \mathcal{N}(0, K_{f})\mathcal{N}(f, \sigma_{n}^{2}\mathbb{I})df \\
&= \mathcal{N}(0, K_{f} + \sigma_{n}^{2}\mathbb{I}) \nonumber
\end{align}
 The matrix $K_{f}$ denotes the kernel $k_{\theta}(\cdot,\cdot)$ evaluated at inputs $X$. 
 
This works considers the non-stationary counterpart of the squared-exponential kernel, the Gibbs kernel \citep{gibbs1998bayesian} for the task of spatial precipitation modelling. In contrast to the SE-ARD kernel proposed above, the Gibbs kernel introduces input dependent lengthscales in each dimension. Instead of a single-point estimate $\ell_{d}$ per dimension we can consider $\ell_{d}(\bm{x})$, a lengthscale function. One can choose a parametric form for the dependence of the lengthscale on the inputs in each dimension but very often this dependence is not known a priori. The Gibbs kernel for multi-dimensional inputs is given by, 
\begin{align}\label{gibbs}
       k_{\text{Gibbs}}(\bm{x}_{i},\bm{x}_{j}) = & \prod_{d=1}^{D}\sqrt{\dfrac{2\ell_{d}(\bm{x}_{i})\ell_{d}(\bm{x}_{j})}{\ell_{d}^{2}(\bm{x}_{i}) + \ell_{d}^{2}(\bm{x}_{j})}} \hspace{3mm} \times 
       \\ & \exp \left\{ - \sum_{d=1}^{D} \dfrac{(x_{i}^{(d)} - x_{j}^{(d)})^{2}}{\ell_{d}^{2}(\bm{x}_{i}) + \ell_{d}^{2}(\bm{x}_{j})}\right \} \nonumber
\end{align}
Previous work, \citet{heinonen2016non} considers a formulation where the lengthscale function is modelled non-parameterically with a latent Gaussian process defined on the same inputs. 
\begin{align}
   \hat{\ell} &= \log(\ell_{d}) \sim \mathcal{GP}(0, k_{\ell}(.,.))\\
      \hat{\ell}(\bm{x}) &= \log(\ell_{d}(\bm{x})) \sim \mathcal{N}(0, K_{\ell})
\end{align}
The hierarchical formulation makes the posterior and marginal likelihood analytically intractable. However, in the Gaussian likelihood setting one can consider a maximum-a-posteriori (MAP) solution \citep{kersting2007most, heinonen2016non} by maximising, 
\begin{align}
    \ell_{\text{MAP}} &= \argmax_{\ell}\log p(\bm{y}|\hat{\ell})p(\hat{\ell}) \\
    &= \argmax_{\ell} \log \mathcal{N}(\bm{y}|0, K_{f} + \sigma^{2}_{n}\mathbb{I})\mathcal{N}(\hat{\ell}|0, K_{\ell}) \nonumber
\end{align}
where $\ell_{\text{MAP}}$ is a vector of the size of the training inputs $X$ (note that it is typical to work with the  log of the marginal likelihood to avoid numerical underflow). Extrapolating the non-parametric lengthscale to test inputs entails estimating $p(\ell_{\star}|\ell_{\text{MAP}}, \bm{y})$ which we approximate by $\mathbb{E}(\ell_{\star}|\ell_{\text{MAP}}) = K_{\star\ell}K_{\ell}^{-1}\ell_{\text{MAP}}$ (expectation of a conditional Gaussian).
\begin{figure*}
    \centering
    \includegraphics[scale=0.6]{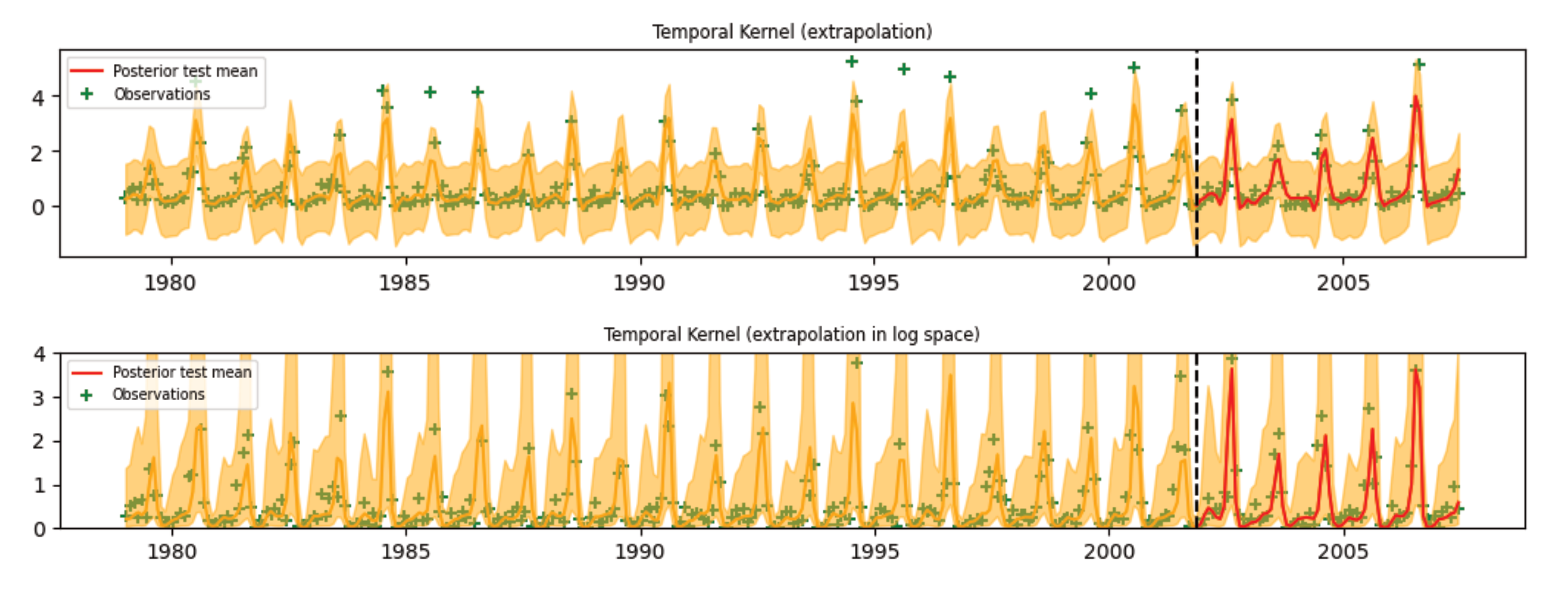}
    \caption{Predicting the precipitation trend with a locally periodic kernel. The dotted line demarcates the training and test regimes. The error bars denote 95$\%$ confidence intervals around the predicted mean. Some outliers remain uncaptured by the Gaussian prediction intervals, indicating mild non-Gaussianity in the targets. In the bottom plot we regress the outputs in log-space and visualise the 1d log-normal distribution with the respective quantiles.}
    \label{fig:temporal}
\end{figure*}
Once the MAP solution has been learnt at the training inputs, the posterior predictive  $p(\bm{f}^{\star}|\bm{y},\ell_{\text{MAP}}, \ell_{\star}) \sim \mathcal{N}(\mu^{\star}, \Sigma^{\star})$ is given by,
\begin{align}
    \mu^{\star} &= K_{\star f}^{T}(K_{f} + \sigma^{2}_{n}\mathbb{I})^{-1}\bm{y} \nonumber \\
    \Sigma^{\star} &= K_{\star\star} - K_{\star f}^{T}(K_{f} + \sigma^{2}_{n}\mathbb{I})^{-1}K_{f\star}
\end{align}
where $K_{f}$ is based on the evaluation of \cref{gibbs} on the training inputs (using $\ell_{\text{MAP}}$) and $K_{\star f}$ is based on the evaluation of $k_{\text{Gibbs}}$ on test and training inputs using $\ell_{\text{MAP}}$ and $\mathbb{E}(\ell_{\star}|\ell_{\text{MAP}})$ respectively. 

The latent GP parameterisation allows for extremely flexible modelling where samples from the posterior function space `adapt' to the varying spatial dynamics inherent in the data.

\section{Case-study: Accurate precipitation modelling in the UIB}
\label{cs}

We conduct three experiments to highlight different features of the kernel composition. In the spatial regression task, we benchmark the non-stationary spatial covariance kernel based on the Gibbs construction against stationary baselines. The temporal extrapolation task uses a locally periodic kernel to fit the univariate precipitation dynamics at single spatial points over time. The spatio-temporal task considers an additive kernel with components acting on the spatial and temporal slice of the inputs to model dynamics over space and time. The models are implemented using \texttt{GPyTorch} \citep{gardner2018gpytorch} and data is plotted using \texttt{xarray} \citep{hoyer2017xarray}. \footnote{Code available at: \newline
\url{https://github.com/kenzaxtazi/climate-kernel-learning}.}

\textbf{Data:} in this case study, we use the 5th ECMWF Reanalysis (ERA5) dataset \citep{hersbach2020era5}. Reanalysis blends historical observations from surface, sonde and satellite measurements with numerical weather forecasting models. Through this data assimilation, the models create a past record of historical climate at high temporal and spatial resolution. ERA5 runs from 1959 to the present day over 0.25° grid and assimilates data from a large number of sources. Each datapoint represents the average monthly precipitation in mm/day over a gridbox.

Precipitation in the UIB is complex and this is reflected in the ERA5 dataset. In this area, precipitation is driven by two major atmospheric events: the Indian Summer Monsoon (ISM) and the western fronts (Westerlies). The ISM brings rain from June to September. The ISM reaches the south-eastern UIB first and has a decreasing contribution to the annual rainfall in the North-West direction. 

The Westerlies are strongest in the winter from December to April, peaking in March \citep{dahri2016appraisal}. The relative contribution of westerly fronts increases from the South-East to the North-West of the basin. As a consequence, the eastern UIB receives up to 70\% of its annual rainfall from the summer whereas the western UIB receives 40-60\% of its precipitation during the winter \citep{dahri2016appraisal}.

The complexity of this distribution is illustrated in Figure \ref{fig:map_and_regimes} and \ref{fig:typical} where three timeseries are sampled from the three characteristic precipitation regimes identified through K-means clustering. 
\begin{figure*}[t]
    \centering
    \includegraphics[scale=0.65]{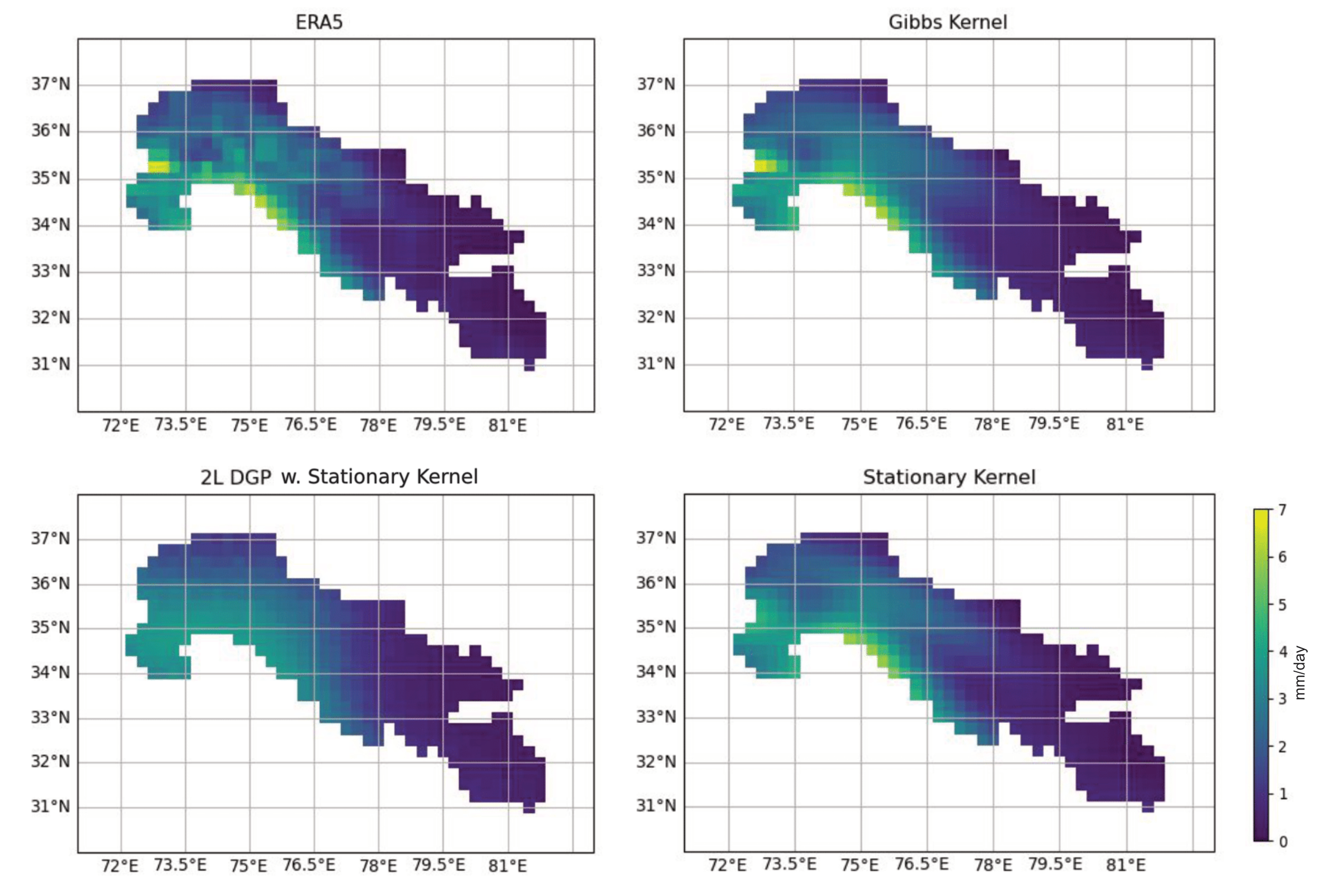}
    \caption{ERA5 data and 2D precipitation modelling over inputs (lat, lon) for the non-stationary Gibbs kernel, and two-layer DGP and shallow GP with stationary kernels. \Cref{tab:spatial} provides test predictive performance metrics over 10 splits of the data.}
    \label{fig:map}
\end{figure*}

\subsection{Spatial Regression}
\label{sr}
We model precipitation in the UIB at a single point in time focusing on accurately capturing the spatial variation of precipitation. The stationary model and the deep GP uses a 2D SE-ARD kernel learning stationary hyperparameters (with ML-II inference) which are constant across space. In the Gibbs formulation (\cref{gibbs}) we learn a lengthscale per spatial point using MAP estimation ( as described in \cref{back}). \Cref{fig:map} shows the best mean prediction from 10 splits for each of the baselines. The 2-layer deep GP \citep{damianou2013deep} with skip connections performs worse than the standard baseline in terms of prediction error. The high-precipitation areas in the Khyber zone are best captured by the Gibbs kernel. It is interesting to note that the predictive intervals for the DGP yield higher predictive densities for test data despite inferior predictive means. The Gibbs formulation gives the best trade-off in terms of reconstruction error assessed against ground-truth and quality of predictive uncertainty.

\begin{table}[h]
\centering
\scalebox{0.9}{
\begin{tabular}{lccc}
\toprule
Metric & SE-ARD & DGP (L2) & Gibbs \\
\midrule
RMSE & 0.353 ± 0.014 & 0.603 ± 0.018  & \textbf{0.271 ± 0.014}\\
NLPD &  0.406 ± 0.063 & 0.023 ± 0.011  & \textbf{0.019 ± 0.016}\\
\bottomrule
\end{tabular}}
\caption{2D Spatial regression on precipitation data over the entire UIB over the month of Jan, 2000. We report Root Mean Squared Error (RMSE) and Negative Log Predictive Density (NLPD) on held-out test data over 10 splits.}
\label{tab:spatial}
\end{table}

\subsection{Temporal Extrapolation}
We regress on univariate time inputs (monthly observations) to capture the local periodicity exhibited in the precipitation dynamics. In order to moderate the changing amplitudes of the peaks we add flexibility by multiplying with a SE-ARD kernel (acting on univariate time inputs). The periodic kernel we use is given by,  
\begin{equation}
    k_{\text{PER}}(\bm{x}_{i}, \bm{x}_{j}) = \sigma^{2}_{f}\exp\left\{ -\dfrac{2\sin^{2}({\pi|\bm{x}_{i} - \bm{x}_{j}|/p})}{\ell^{2}}\right\}
\end{equation}

\Cref{fig:temporal} depicts the prediction performance over train and test where we train on years 1979-2002. The predictions under the log-normal distribution ensure positivity (as desired), further the heavy-tailed non-Gaussian prediction intervals capture the outlying values which elude the symmetric Gaussian intervals. The test RMSE and NLPD for the fits are given by (0.5536 vs. 0.5328) and (1.3248 vs. 0.9721).

\subsection{Spatio-Temporal Regression}

In this experiment we consider the task of predicting dynamics across space and time (3D inputs) for a 1 month ahead forecast. We train on January to April 2000 and test on the month of May. We wanted to constrain the training data set to a moderate size so as to execute exact GP inference for the baseline case.

For the stationary (shallow) model, the spatio-temporal kernel is formed by adding together the spatial and temporal components with one important innovation. In the earlier sections the temporal component acted solely on the time dimension while the spatial component acted solely on the spatial dimensions (latitude and longitude). Denoting the time and spatial coordinates as $\bm{x}^{\text{t}}, \bm{x}^{\text{lat}}, \bm{x}^{\text{lon}}$ respectively, the kernel for the stationary model is given by,
\vspace{-2mm}

\begin{align}
& k_{\text{stat.}}(\bm{x}_{i}, \bm{x_{j}}) = \nonumber \\ 
&\underbrace{k_{\text{SE-ARD}}(({\bm{x}^{\text{lat}}_{i}},{\bm{x}^{\text{lon}}_{i}}), ({\bm{x}^{\text{lat}}_{j}},{\bm{x}^{\text{lon}}_{j}}))\times k_{\text{PER}}({\bm{x}^{t}_{i}}, {\bm{x}^{t}_{j}})}_{\text{temporal}}\nonumber \\ & +  \underbrace{k_{\text{SE-ARD}}(({\bm{x}^{\text{lat}}_{i}},{\bm{x}^{\text{lon}}_{i}}),({\bm{x}^{\text{lat}}_{j}},{\bm{x}^{\text{lon}}_{j}}))}_{\text{spatial}}
\label{kstat}
\end{align}

\begin{figure*}[h]
    \centering
    \includegraphics[width=\textwidth]{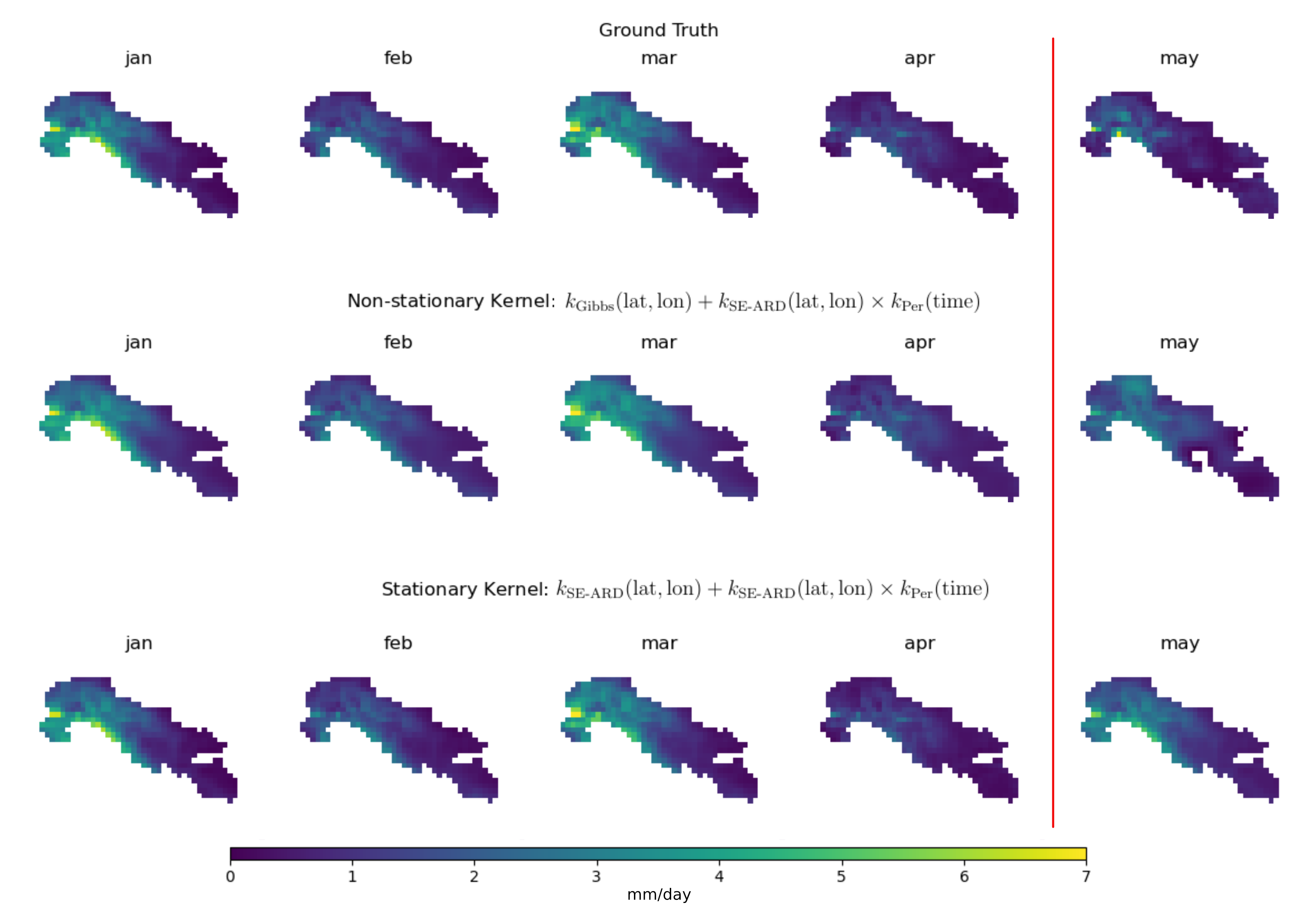}
    \caption{3D regression for spatio-temporal precipitation modelling. The red line demarcates the training and test months. The non-stationary kernel yielded a an average test RMSE of 0.9426 across three runs vs. 1.1086 for the stationary kernel.}
    \label{fig:spatio_temporal}
\end{figure*}

We weave the spatial dimensions into the temporal kernel by making the RBF component dependent on the latitude and longitude, this allows the amplitude/scale of the periodic component to vary for different spatial regions. Note that this is still a stationary kernel. For the non-stationary extension in this setting we keep the temporal component identical to the stationary formulation \cref{kstat} but replace the spatial component with the Gibbs kernel \cref{gibbs}.
\vspace{-3mm}
\begin{align}
& k_{\text{non-stat.}}(\bm{x}_{i}, \bm{x_{j}}) = \nonumber \\ 
&\underbrace{k_{\text{SE-ARD}}(({\bm{x}^{\text{lat}}_{i}},{\bm{x}^{\text{lon}}_{i}}), ({\bm{x}^{\text{lat}}_{j}},{\bm{x}^{\text{lon}}_{j}}))\times k_{\text{PER}}({\bm{x}^{t}_{i}}, {\bm{x}^{t}_{j}})}_{\text{temporal}}\nonumber \\ & +  \underbrace{k_{\text{Gibbs}}(({\bm{x}^{\text{lat}}_{i}},{\bm{x}^{\text{lon}}_{i}}),({\bm{x}^{\text{lat}}_{j}},{\bm{x}^{\text{lon}}_{j}}))}_{\text{spatial}}
\end{align}
where $k_{\text{Gibbs}}$ is parameterised by 2 non-parametric lengthscale processes - $\ell_{\text{lat}}({\bm{x}^{\text{spatial}}})$ and $\ell_{\text{lon}}({\bm{x}^{\text{spatial}}})$ acting on the 2d spatial dimensions as inputs. We enclose the regression results for the 1-month ahead forecast based on the baseline and non-stationary construction in \cref{fig:spatio_temporal}.

We do not present results for the deep GP in this sections as they were found to yield substandard results in \cref{sr}.

\section{Conclusions and further work}\label{sec:conclusion}

We demonstrate kernel composition in modelling the spatial and temporal precipitation dynamics over the UIB. In the spatial setting, the non-stationary kernel does much better than a deep GP and the stationary baseline. More comprehensive and complex experiments will provide answers to how these methods can be best be used in `real world' settings and, scientifically, what drives precipitation in the UIB.

Further investigation will seek to better understand historical precipitation distribution in this area. In particular, a detailed analysis of the Gibbs hyperparameter functions are needed over longer training and testing intervals. This would require making the non-stationary formulation compatible with sparse GPs relying on inducing points \citep{titsias2009variational}. Future iterations will also consider more features that are predictive of precipitation such as elevation, slope and large scale atmospheric variables. The Gibbs hyperparamaters will give us insight into where and how these features are influencing rain and snowfall in basin. Model comparison metrics like BIC (Bayesian information criterion) will help understand more complex non-stationary constructions in light of model complexity.

Next, precipitation change under different climate scenarios will be considered. ERA5 climatic variables will be swapped with global climate model outputs. Although not explored in this paper, the non-stationarity of precipitation with respect to climate change will become more important in the future. This non-stationarity should be in part captured by the large scale atmospheric variables previously mentioned. However, a non-stationary kernel could also have a role to play here. 

Finally, the Bayesian nature of these models will give a more accurate and informed quantification of uncertainty than regional climate models that uses ensemble spread as proxy for uncertainty. In the non-stationary setup, the hyperaparemeter distributions could also point scientists to the most significant drivers of prediction variability.

\bibliography{main}

\providecommand{\upGamma}{\Gamma}
\providecommand{\uppi}{\pi}

\end{document}